\documentclass[10pt,twocolumn,letterpaper]{article}

\usepackage[pagenumbers]{iccv} 

\usepackage[dvipsnames]{xcolor}

\definecolor{iccvblue}{rgb}{0.21,0.49,0.74}
\usepackage[pagebackref,breaklinks,colorlinks,allcolors=iccvblue]{hyperref}

\usepackage[ut  f8]{inputenc} 
\usepackage[T1]{fontenc}    
\usepackage[pagebackref]{hyperref}       
\usepackage{url}            
\usepackage{booktabs}       
\usepackage{amsfonts}       
\usepackage{nicefrac}       
\usepackage{microtype}      
\usepackage{xcolor}         
\usepackage{nicematrix}

\usepackage{enumitem}
\usepackage{amsmath}
\usepackage{multirow}
\usepackage{graphicx}
\usepackage{tabularray}
\usepackage{pifont}
\usepackage{graphicx}
\newcommand{\davg}{\delta_{\textrm{avg}}}
\usepackage{bbm}
\usepackage[accsupp]{axessibility}  
\usepackage{colortbl}

\def\paperID{4753} 
\def\confName{ICCV}
\def\confYear{2025}

 \title{AllTracker: Efficient Dense Point Tracking at High Resolution}

\author{%
Adam W. Harley\textsuperscript{1}\quad Yang You\textsuperscript{1}\quad Xinglong Sun\textsuperscript{1}\quad Yang Zheng\textsuperscript{1}\quad Nikhil Raghuraman\textsuperscript{1} \and Yunqi Gu\textsuperscript{1} \quad Sheldon Liang\textsuperscript{2}\quad Wen-Hsuan Chu\textsuperscript{2}\quad Achal Dave\textsuperscript{3}\quad  Pavel Tokmakov\textsuperscript{3}  \and Suya You\textsuperscript{4} \quad Rares Ambrus\textsuperscript{3}\quad Katerina Fragkiadaki\textsuperscript{2}\quad Leonidas Guibas\textsuperscript{1}\\ 
\small{\textsuperscript{1}Stanford University \quad \textsuperscript{2}Carnegie Mellon University \quad\textsuperscript{3}Toyota Research Institute \quad \textsuperscript{4}Army Research Laboratory}
}

\renewcommand{\paragraph}[1]{\vspace{0.25em}\noindent\textbf{#1}\quad}

\begin{document}

\maketitle

\begin{abstract}

We introduce AllTracker: a model that estimates long-range point tracks by way of estimating the flow field between a query frame and every other frame of a video. Unlike existing point tracking methods, our approach delivers high-resolution and dense (all-pixel) correspondence fields, which can be visualized as flow maps. Unlike existing optical flow methods, our approach corresponds one frame to hundreds of subsequent frames, rather than just the next frame.  We develop a new architecture for this task, blending techniques from existing work in optical flow and point tracking: the model performs iterative inference on low-resolution grids of correspondence estimates, propagating information spatially via 2D convolution layers, and propagating information temporally via pixel-aligned attention layers. The model is fast and parameter-efficient (16 million parameters), and delivers state-of-the-art point tracking accuracy at high resolution (i.e., tracking $768 \times 1024$ pixels, on a 40G GPU). A benefit of our design is that we can train jointly on optical flow datasets and point tracking datasets, and we find that doing so is crucial for top performance. We provide an extensive ablation study on our architecture details and training recipe, making it clear which details matter most. Our code and model weights are available: \url{https://alltracker.github.io}

\end{abstract}

\begin{figure}[ht]
    \centering
    \includegraphics[width=1.0\linewidth]{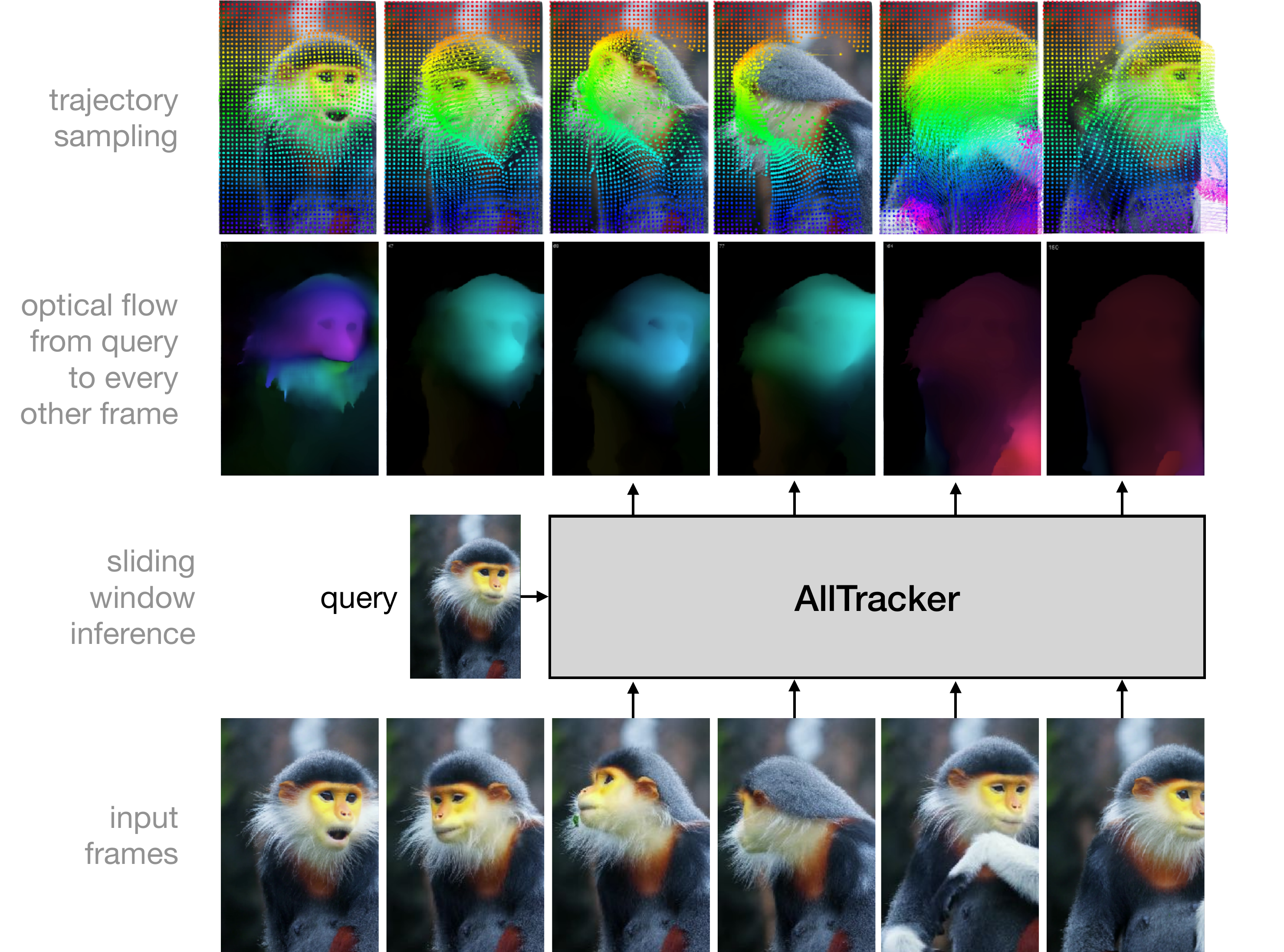}
    \caption{\textbf{AllTracker} estimates high-resolution optical flow between a ``query frame'' and every other frame of a video, using a sliding-window strategy. 
    Point samples from these outputs can be interpreted as long-term point trajectories.}
    \label{fig:teaser}
\end{figure}

\section{Introduction}\label{sec:intro}

Estimating the long-range trajectories of arbitrary points in the world, using sequences of 2D images as input, has been a concrete and competitive challenge in computer vision since at least 2006~\cite{particlevideo}. 
The utility of optical flow (i.e.,  the instantaneous velocity of pixels~\cite{gibson1950perception}) toward this goal has long been obvious, yet it has remained challenging to upgrade flows into long-range tracks.

Instantaneous flows can be ``chained'' into multi-frame tracks, by interpolating in one flow field at the endpoints of the previous flow field, but imperfect flows will accumulate drift, and such chains must also be carefully stopped at occlusions~\cite{particlevideo}. 
An attractive shortcut here is to directly compute the flow between a reference frame and each other frame, and thus track drift-free and across occlusions, 
but estimating ``long-range flow'' becomes increasingly difficult as the time interval widens between the reference frame and the target, due to increasing variation in perspective, illumination, and  scene geometry. 

In light of these challenges, recent work has 
side-stepped optical flow and created 
a line of bespoke point trackers focused on learning multi-frame temporal priors for 
reducing drift and tracking through occlusions~\cite{harley2022particle,zheng2023point,karaev2023cotracker,doersch2023tapir, karaev2024cotracker3}.   
These works explicitly highlight that the limited temporal context in flow-based methods is a severe weakness, and have made substantial progress 
by addressing this. However, the proposed solutions add temporal awareness at the cost of spatial awareness, and only deliver tracks for sparse sets of points. Recent attempts at ``dense'' point tracking~\cite{le2024dense,Tournadre_2024_DTF,ngo2024delta} are not as accurate as the latest sparse trackers, and struggle with high-resolution input. 

In this paper, we demonstrate that learnable multi-frame temporal priors can 
be built together with high-resolution spatial awareness, 
by casting point tracking as a \textit{multi-frame} long-range optical flow problem, as illustrated in Figure~\ref{fig:teaser}.  

Our design combines ideas from point tracking literature and optical flow literature, arriving at a composite architecture with broader capability. Following the trend in both areas, the heart of the model is a recurrent module that iteratively improves motion estimates~\cite{raft,harley2022particle}. %
This module relies on information from spatial cross-correlations, 
giving it a strong inductive bias for feature-matching, which speeds up training. 
From point tracking literature, we borrow a per-pixel temporal module, to learn a motion prior and track through occlusions~\cite{harley2022particle}. 
From optical flow literature, we borrow the idea of performing the majority of processing on a low-resolution grid, to allow fast spatial message-passing via 2D convolutions, and recover spatial precision at the end of the architecture with an upsampling layer~\cite{raft, wang2024sea}. 
In relation to other point trackers, the key novelty of our approach is that we frame the problem as long-range dense flow, which makes sparse methods (of similar speed and accuracy) redundant. 
In relation to other optical flow models, the key novelty of our approach is that we solve a \textit{window} of flow problems simultaneously, instead of frame-by-frame; 
the information shared within and across windows unlocks the capability to resolve flows across wide time intervals. 

A secondary benefit of our design is that we can train jointly on optical flow datasets and point tracking datasets. We therefore build a mix of publicly-available datasets to support our model, spanning different time-ranges, annotation densities, and resolutions, and train with uniform sampling across this mix. We find that this strategy, paired with a long training schedule, is crucial for top performance. 

In summary, our method 
blends techniques and datasets from optical flow literature and point tracking literature, 
and the resulting model is a state-of-the-art point tracker that operates in high resolution at full density. Inspired by CoTracker~\cite{karaev2023cotracker} and highlighting our capability for all-pixel tracking, we name our approach \textit{AllTracker}.  
We will release our model, our data, and our code.

\section{Related  Work} \label{sec:related}

\paragraph{Optical flow}
The introduction of the concept of optic flow can attributed to Gibson~\cite{gibson1950perception} (see Niehorster's review~\cite{niehorster2021optic}). In computer vision, optical flow methods generally take two consecutive frames of video as input, and estimate the motion that relates the pixels of the first frame to the pixels of the second frame~\cite{horn1981determining, lucas1981iterative}. 
The dominant technique, from classic approaches to current learning-based approaches, is to iterate on a solution: first, candidate correspondences are initialized (e.g., with zero-motion or constant velocity), then the ``costs'' of these correspondences are computed by measuring how well the appearance features match, and this appearance cost is combined with smoothness terms or learned priors, and then better correspondences are sought in the neighborhood of the current solution~\cite{brox2010large,ilg2017flownet,xu2017accurate,sun2018pwc,raft}. Our work uses the same basic technique. 

The current state of the art in optical flow estimation is  SEA-RAFT~\cite{wang2024sea}. 
SEA-RAFT first directly estimates a low-resolution flow field~\cite{flownet,ilg2017flownet}, then computes appearance features with a ResNet-34~\cite{he2016deep}, and computes pixel-to-pixel correlations with these features~\cite{xu2017accurate}, then iteratively estimates refinements to the flow via 2D convolutions that take flows and correlations as input, and finally upsamples the flow to full resolution via a pixel-shuffle layer~\cite{shi2016real}. 
Our work takes inspiration from SEA-RAFT (and its predecessor RAFT~\cite{raft}) in the way that we iteratively refine low-resolution flows and upsample to full resolution, but we operate on multiple frames at a time (i.e., 16 frames instead of 2), and we add temporal attention layers to help information propagate across the time axis. 

\paragraph{Flow-based point tracking}
Optical flow is often used as a building block toward multi-frame tracking. The standard technique in this space is to ``chain'' flow vectors end-to-end to form longer tracks, and guard against drift by monitoring for occlusions~\cite{particlevideo,brox_densepoint}. Recent techniques skip past occlusions by using multistep flows~\cite{crivelli2012optical, conze2014dense, conze2016multi, neoral2023mft}, or attempt to track through occlusions using a learned temporal prior~\cite{cho2024flowtrack} (inspired by non-flow point trackers). These methods deliver dense multi-frame trajectories similar to our method, but first require an expensive pre-processing stage in which optical flows are estimated by a different model, while our method handles the full problem efficiently with a single model. 

We highlight two concurrent works which propose a flow-based method somewhat similar to our own: DTF~\cite{Tournadre_2024_DTF} and DELTA~\cite{ngo2024delta}. 
Both iteratively estimate flows to relate a reference frame to other frames, 
but unlike our work, these papers propose special-purpose transformer-based methods in which global spatial message-passing is approximated by cross-attending to sparse ``anchor'' or ``centroid'' tokens. Our method simply uses 2D convolutions for this step, more like SEA-RAFT~\cite{wang2024sea}. 
We note also that DTF and DELTA both struggle with high-resolution inputs (running out of memory on our 40G GPUs), despite DELTA using low-resolution inference and high-resolution upsampling, comparable to SEA-RAFT and our approach. 
In performance, DTF is not competitive with recent point trackers (while ours exceeds them); DELTA's main model requires depth input, but we compare against its 2D variant and outperform it. 

\paragraph{Point trackers without flow}
Many classic methods track points directly, without relying on optical flow as a submodule~\cite{lucas1981iterative,tomasi1991detection}. 
Harley et al.~\cite{harley2022particle} recently introduced a new method in this space, 
with some components similar to flow models~\cite{raft}, but tracking points independently, and adding a multi-frame inference window which enabled the model to track through occlusions. That work, along with a new ``Tracking Any Point'' benchmark that appeared soon afterward~\cite{doersch2022tap}, spurred much follow-up effort, involving multi-point context~\cite{karaev2023cotracker,karaev2024cotracker3}, better initialization schemes~\cite{doersch2023tapir}, wider correlation context~\cite{bian2024context,cho2025local}, new transformer-based designs~\cite{li2024taptr,li2024taptrv2}, and post-hoc densification methods~\cite{le2024dense}. 

The current state of the art in point tracking is  CoTracker3~\cite{karaev2024cotracker3}. CoTracker3 first computes feature maps for all frames, then initializes tracks for a sparse set of query points, then retrieves local correlations based on these tracks, and interleaves modules which pass information within tracks (temporally) and across tracks (spatially). The main novelty is in the spatial propagation: each point attends to a set of latent ``virtual points'' which learn a compressed representation of the video motions. 
While CoTracker3 delivers very accurate tracks, it requires users to pre-select a sparse set of points to track, and results vary based on how the queries are distributed~\cite{karaev2023cotracker}. CoTracker3 is more efficient than its predecessors~\cite{karaev2023cotracker,harley2022particle}, but is still limited to tracking a few thousand points at a time. 
Our approach is simpler and more memory-efficient: we track on a low-resolution grid, where we achieve message-passing with simple 2D convolutions, and we recover spatial precision at the end of the architecture with a fast upsampling layer~\cite{shi2016real}. 
We note that this efficient upsampling technique was known well before the recent resurgence of point trackers~\cite{shi2016real}, but perhaps its relevance was not fully appreciated.

\paragraph{Training data and self-supervision}
Current methods in optical flow and point tracking are data-driven, making data and supervision choices crucial. In optical flow, state-of-the-art methods rely on a combination of mostly synthetic datasets, 
including FlyingChairs \cite{flownet}, FlyingThings3D~\cite{mayer2016large},  Monkaa~\cite{mayer2016large}, Driving~\cite{mayer2016large}, AutoFlow~\cite{sun2021autoflow}, Spring~\cite{Mehl2023_Spring}, VIPER~\cite{richter2017playing}, HD1K~\cite{kondermann2016hci}, KITTI~\cite{kitti}, and TartanAir~\cite{wang2020tartanair}. In point tracking, the main training datasets are Kubric~\cite{greff2022kubric}, FlyingThings++~\cite{harley2022particle}, and PointOdyssey~\cite{zheng2023point}. 
There is an art to creating a curriculum from diverse datasets 
to optimize downstream performance on particular benchmarks~\cite{wang2024sea}, but in our work we simply concatenate the datasets and shuffle the samples. We note that all previous point tracking methods did not use flow data, perhaps because most models do not support 2-frame inference or dense output; we therefore also provide experiments using Kubric alone. 

Motivated by the fact that the training data for point trackers is all synthetic, several groups have explored self-supervision methods based on boot-strapping pre-trained models using pseudo-labels computed on real videos~\cite{sun2024refining,doersch2024bootstap,karaev2024cotracker3}. 
However, these schemes offer only minor gains over the supervised weights where bootstrapping begins. In this work, we avoid any pseudo-labelling and simply add more synthetic data, capitalizing on our model's ability to accept supervision from optical flow data.

\section{AllTracker}\label{sec:method}

\begin{figure}[t]
\includegraphics[width=1.0\linewidth]{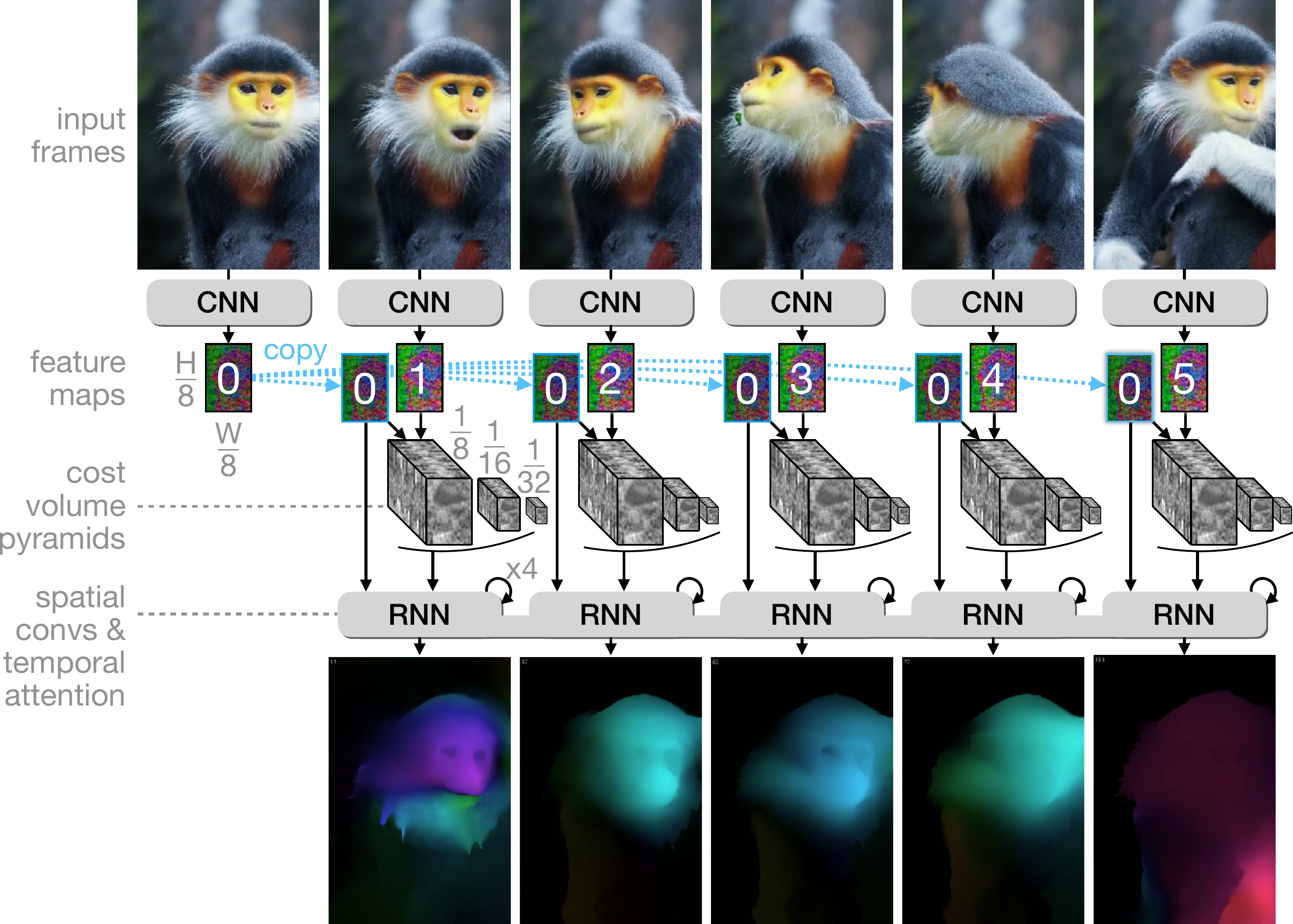}
 \caption{\textbf{AllTracker architecture.} 
First, we compute feature maps for all frames, and copy the zeroth (query) feature map to every timestep, and compute multi-scale cost volumes. 
Then, we iterate a recurrent module, which references the query feature map and cost volume pyramid at each timestep, and estimates a low-resolution correspondence field, using interleaved 2D convolutions and pixel-aligned temporal attentions. 
The output of the RNN is upsampled into high-resolution optical flow maps, which relate all pixels of the zeroth frame to every other frame. 
 } 
 \label{fig:arch}
\end{figure}

Our model takes a video as input, along with a ``query'' index, specifying which frame's pixels to track, and it outputs full-timespan tracking for all of the pixels in that frame. 

Concretely, we are given as input a video of shape $T,H,W,3$, where $T$ is the number of frames, and $H,W$ indicate the spatial resolution of each frame. We are also given an index $t \in T$, 
indicating which frame has the pixels we need to track. Our final output is a tensor shaped $T,H,W,4$: the first two channels are optical flow maps indicating the offset that takes each pixel in the query frame to its correspondence in every other frame, and the next two channels estimate visibility and confidence. 

A brief summary of the method is as follows. We operate in sliding window fashion across $T$, processing subsequences of length $S$, advancing at a stride of $S/2$. 
We begin by quickly computing low-resolution feature maps for the subsequence, yielding a feature volume with dimensions $S, H/8, W/8, D$, where $D$ is the channel dimension of the features. 
We then initialize a low-resolution tensor of outputs, shaped $S, H/8, W/8, 4$, and iteratively revise this tensor 
using cross correlations and features across space and time as reference. 
Finally, we upsample the outputs, producing full resolution optical flow, visibility, and confidence, shaped $S,H,W,4$. 
We then advance our window by $S/2$, using previous estimates as initialization, 
and repeat inference. 
Figure~\ref{fig:arch} shows the method in a diagram.

\subsection{Encoding}\label{sec:enc}

Our first stage encodes the input video frames into low-resolution feature maps. We achieve this with a ConvNeXt-Tiny~\cite{liu2022convnet} model (see Sec.~\ref{sec:impl} for implementation details). 
This encoder compresses a subsequence input of shape $S,H,W,3$ to the shape $S,H/8,W/8,D$, where $D$ is the embedding dimension (in our case $D=256$), $S$ is the subsequence window length (in our case $S=16$). When the full video length $T$ is not too large (e.g., at training time), we compute all of the feature maps in parallel. 

We then isolate the feature map of the {query} frame and tile it to the length of the subsequence, so that we have a copy for each timestep, as illustrated in Figure~\ref{fig:arch}.

\subsection{Computing appearance similarity}
Using the feature maps, we build a multi-scale 4D correlation volume, capturing appearance-based tracking cues. 
We implement this in two steps: first we convert each feature map into a feature pyramid, by average-pooling at multiple strides (in our case 5 strides: $\{1, 2, 4, 8, 16\}$), and then we cross-correlate the query feature map with these pyramids. That is, for each feature vector in the query feature map, we compute its dot product across each timestep's pyramid.

The output of this step can be understood as a large collection of heatmaps: one heatmap per tracked pixel, per scale, per timestep. 
The model will index into these heatmaps to retrieve 
information on where correspondences lie. When a target's correspondence is visible at a certain timestep, we expect the respective heatmap to have a strong peak at the location of the correspondence. 

\subsection{Track initialization}

We next initialize estimates for tracking coordinates, visibility, and confidence, by creating a tensor with shape $S,H/8,W/8,4$, where the coordinate channels are initialized with a 2D meshgrid, and visibility and confidence are initialized with zeros. If we are on a window beyond the first, we populate this tensor with the estimates that overlap, and copy forward the values from the last estimated timestep.

\subsection{Iterative refinement}

The most important stage of our model is the iterative refinement stage, where we update all of the estimates. We describe the main points of the module here, but also provide extensive details (and a diagram) in the supplementary.

For each timestep, for each pixel in the query frame, we have: a feature vector $\mathbf{f}$ (length $D$), visibility and confidence estimates $\mathbf{v,c}$ (together length 2), %
a position estimate $\mathbf{p}$ (length $2$), and a correlation pyramid $\{\mathbf{C}_1, \mathbf{C}_2, \ldots \}$ shaped $\{ H/8 \times W/8, H/16 \times W/16, \ldots \}$. 
We convert this data into a ``local'' representation suitable for convolutions, as follows. 
For the position data, we replace the absolute position estimates $\mathbf{p}$ with motion estimates, by subtracting the source positions: $\mathbf{m} = \mathbf{p} - \mathbf{p}_0$. 
For each level of the correlation pyramid, we extract a small patch centered at $\mathbf{p}$, and flatten these patches into a single vector $\mathbf{q}$ of length $L \cdot (2R+1)^2$, where $R$ is a radius parameter for the patches (e.g., 4) and $L$ is the number of levels in the pyramid (e.g., 5). All together, we have 
$\mathbf{f}, \mathbf{v}, \mathbf{c}, \mathbf{m}, \mathbf{q}$ (per pixel, per timestep), which in our setup makes $665$ channels.

Inside the recurrent module, we process the input data using interleaved spatial and temporal blocks. Each spatial block is a 2D ConvNeXt block; each temporal block is a \textit{pixel-aligned} transformer block. By pixel-aligned, we mean that attention only happens along the temporal axis (i.e., with cost quadratic in $S=16$), and this is done for every pixel in parallel. 
We note that an important convenience of our design is that all of the tensors in this stage are aligned with the ``query'' frame; therefore pixel-aligned attention is attention between \textit{corresponding} pixels. After the interleaved spatial and temporal blocks propagate tracking-related information across our window, we decode explicit revisions for visibility, confidence, and motion.

We apply visibility, confidence, and motion revisions via simple summation with the previous values 
$\mathbf{x}_\textrm{new} = \mathbf{x}_\textrm{old} + \delta \mathbf{x}$ for $\mathbf{x} \in \{\mathbf{v}, \mathbf{c}, \mathbf{m} \}$, 
where $\delta \mathbf{x}$ denotes an explicit revision produced by the model.

We additionally decode weights for a pixel-shuffle upsampling step~\cite{shi2016real,raft}. We apply this upsampling to our visibility, confidence, and motion maps, bringing them from $1/8$ resolution to full resolution.

We iterate our recurrent ``refinement'' stage 4 times, sharing weights. 
At training time, we use the outputs from every refinement step for supervision, 
and at test time we only use the final iteration's output. 

\subsection{Model training}

In the datasets which we use for training, we have supervision in the form of either optical flow or sparse point tracks, which we simply treat as trajectories of varying lengths. We use an $L_1$ loss between estimated trajectories and ground truth, with a higher weight on loss for visible points: 
\begin{equation}
    L_\textrm{track} = \alpha \sum_k^K  \gamma^{K-k} ( \mathbbm{1}_\textrm{occ}/5 + \mathbbm{1}_\textrm{vis} ) || P_k - \hat{P} ||_1,
\end{equation}
where $P_k$ is a set of estimated trajectories at refinement step $k$, $\hat{P}$ is the corresponding ground truth, $\gamma$ (set to $0.8$) makes later refinement steps weigh more in the loss, and $\alpha$ is a balancing hyperparameter (set to $0.05$). 

We supervise our visibility and confidence maps with a binary cross entropy loss. We ask the visibility estimates to match ground truth binary labels: $L_\textrm{vis} = \sum_k^K \textrm{BCE}(V, \hat{V})$. 
We ask the confidence estimates to reflect whether or not the corresponding position estimates are within $12$ pixels of ground truth~\cite{doersch2023tapir,karaev2024cotracker3}: 
 $   L_\textrm{conf} = \sum_k^K  
    \textrm{BCE}(C, \mathbbm{1}[|| X_k - \hat{X} ||_2 < 12 ])$.
We apply these losses at every refinement step.

    To supervise the model with sparse annotations, we use the coordinates of the ground truth to sample our corresponding estimates, 
and apply the loss at these sparse locations. 
We use bilinear sampling for trajectories, and nearest-neighbor sampling for visibility and confidence.

\subsection{Implementation details}\label{sec:impl}

We provide extensive implementation details in the supplementary material, but describe the main details here. 

Our CNN backbone is based on a pre-trained ConvNeXt-Tiny~\cite{liu2022convnet}. We use the first three ``blocks'' of this architecture, totaling 12.72 million parameters. We convert the third block from stride 2 to stride 1, by applying bicubic interpolation to the $2\times2 $ stride-$2$ kernel to create a $3 \times 3$ stride-1 kernel, and re-scale these weights according to the change in area (i.e., scaling by $4/9$).

Our full model is 16.48 million parameters. We train it using 8 A100-40G GPUs. We train in two stages: first on Kubric~\cite{greff2022kubric} for 200,000 steps at a learning rate of $5e-4$, and then on a mix of point tracking and optical flow datasets for 400,000 iterations at a learning rate of $1e-5$. We use the standard augmentations from prior work in point tracking~\cite{harley2022particle} and optical flow~\cite{raft}, which consist of random shifting and scaling, color jitter, and square occlusions. 

Our optical flow datasets include FlyingChairs \cite{flownet}, FlyingThings3D~\cite{mayer2016large},  Monkaa~\cite{mayer2016large}, Driving~\cite{mayer2016large}, AutoFlow~\cite{sun2021autoflow}, Spring~\cite{Mehl2023_Spring}, VIPER~\cite{richter2017playing}, HD1K~\cite{kondermann2016hci}, KITTI~\cite{kitti}, and TartanAir~\cite{wang2020tartanair}. Our point tracking datasets include Kubric~\cite{greff2022kubric}, DynamicReplica~\cite{karaev2023dynamicstereo},  
and PointOdyssey~\cite{zheng2023point}. We also follow Harley et al.~\cite{harley2022particle} and compute point tracks from optical flows where possible, making FlyingThings++, Monkaa++, Driving++, and Spring++.

For trajectory estimates, we apply the model's additive revisions directly in pixel coordinates. We also supervise in pixel coordinates, which is a reason to scale the loss by the factor $\alpha = 0.05$. 
For visibility and confidence estimates, 
we apply the model's revisions on logits (rather than on probabilities),   
but we apply a sigmoid to these values before they are input to the refinement block, to stabilize their range.

\definecolor{highlight}{HTML}{E6E6E6}

\newcommand{\mycc}{\cellcolor{lightgray}}
\newcommand\colorrow{\rowcolor{lightgray}\cellcolor{white}}

\begin{table*}[t]
\small 
\centering
\caption{Comparison against recent point trackers and optical flow models, across nine datasets.  We evaluate $\delta_\text{avg}$ (higher is better), using an input resolution of $384 \times 512$. The benchmarks are BADJA~\cite{badja}, CroHD~\cite{sundararaman2021tracking}, TAPVid-DAVIS~\cite{doersch2022tap}, DriveTrack~\cite{drivetrack}, EgoPoints~\cite{egopoints}, Horse10~\cite{horse10}, 
TAPVid-Kinetics~\cite{doersch2022tap}, 
RGB-Stacking~\cite{rgbstacking}, and RoboTAP~\cite{vecerik2024robotap}. 
} \vspace{-0.5em}
\begin{tabular}{lccccccccccc>{\columncolor[gray]{0.9}}c}
\toprule
Method & Params. & Training & Bad. & Cro. & Dav. & Dri. & Ego. & Hor. & Kin. & Rgb. & Rob. & {Avg.}\\
\midrule
RAFT~\citep{raft}& 5.26 &Flow mix&23.7&29.3&48.5&44.8&41.0&27.8&64.3&82.8&72.2&{48.3}\\
SEA-RAFT~\citep{wang2024sea} & 19.66 &Flow mix&23.9&21.9&48.7&49.4&44.0&33.1&64.3&85.7&67.6&{48.7}\\
AccFlow~\citep{wu2023accflow} &11.76 &Flow mix&10.3&22.2&23.5&26.4&4.0&12.1&38.8&63.2&57.9&{28.7}\\
PIPs++~\cite{zheng2023point} & 17.57 &PointOdyssey&34.1&27.5&62.5&51.3&38.5&21.4&64.2&70.4&73.4&{49.3}\\
LocoTrack~\cite{cho2025local} & 11.52 &Kubric&41.4&43.1&68.0&66.5&58.4&48.9&70.0&80.3&76.9&{61.5}\\
{BootsTAPIR}~\cite{doersch2024bootstap} & 54.70 & Kubric+15M & 42.7&34.9&67.9&66.9&56.8&48.8&70.6&81.0&78.2&{60.9}\\
DELTA~\cite{ngo2024delta} & 59.17 &Kubric&44.6&42.9&75.3&67.8&40.3&41.8&66.5&83.0&74.8&{59.7}\\
{CoTracker2~\cite{karaev2023cotracker}} & 45.43 & Kubric&40.0&31.7&70.9&67.8&43.2&33.9&65.8&73.4&73&{55.5}\\
{CoTracker3-Kub~\cite{karaev2024cotracker3}} & 25.39 & Kubric&47.5&\textbf{48.9}&\textbf{77.4}&\textbf{69.8}&58.0&47.5&70.6&83.4&77.2&{64.5}\\
{CoTracker3~\cite{karaev2024cotracker3}} & 25.39 & Kubric+15k&48.3&44.5&77.1&\textbf{69.8}&60.4&47.1&71.8&84.2&81.6&{65.0}\\
AllTracker-Tiny-Kub & 6.29&Kubric&45.4&39.6&73.7&65.1&55.9&45.2&70.6&86.1&79.3&{62.3}\\
AllTracker-Tiny & 6.29&Kubric+mix&47.5&39.8&74.3&63.9&58.3&45.5&71.5&88.1&80.7&{63.3}\\
AllTracker-Kub& 16.48
&Kubric&46.4&42.3&75.2&66.1&60.3&\textbf{49.0}&71.3&\textbf{90.1}&82.2&{64.8}\\
AllTracker&16.48&Kubric+mix&\textbf{51.5}&44.0&76.3&65.8&\textbf{62.5}&\textbf{49.0}&\textbf{72.3}&90.0&\textbf{83.4}&{\textbf{66.1}}\\
\bottomrule
\end{tabular}
\label{tab:singleres}
\end{table*}
\begin{table*}[t]
\small
\centering
\caption{High-resolution comparison between our model and CoTracker3~\cite{karaev2024cotracker3}, which is the previous state of the art. We evaluate $\delta_\text{avg}$ (higher is better) on nine point tracking benchmarks, at resolutions $448 \times 768$ and $768 \times 1024$. 
Parameter counts are in millions.
} \vspace{-0.5em}
\begin{tabular}{lccccccccccc>{\columncolor[gray]{0.9}}c}
\toprule
Model & Params. & Resolution & Bad. & Cro. & Dav. & Dri. & Ego. & Hor. & Kin. & Rgb. & Rob. & {Avg.}\\

\midrule
{CoTracker3-Kub~\cite{karaev2024cotracker3}} & 25.39 & \small{$448\times768$}&49.7&59.9&78.8&70.2&57.4&50.6&70.4&81.1&76.6&{66.1}\\
{CoTracker3~\cite{karaev2024cotracker3}} & {25.39} & \small{$448\times768$}&50.7&\textbf{57.9}&\textbf{79.5}&\textbf{70.8}&61.3&51.1&72.3&82.8&81.3&{67.5}\\
{AllTracker-Tiny-Kub} & {6.29} & \small{$448\times768$}&47.7&48.0&76.3&67.8&57.3&48.3&71.2&84.8&79.0&{64.5}\\
{AllTracker-Tiny} & {6.29} & \small{$448\times768$}&49.7&48.8&77.0&67.8&60.7&49.2&72.3&88.1&80.0&{66.0}\\
{AllTracker-Kub} & {16.48} & \small{$448\times768$}&49.8&51.6&77.6&68.3&61.3&\textbf{52.6}&71.9&90.4&82.0&{67.3}\\
{AllTracker} & {16.48} & \small{$448\times768$}&\textbf{52.5}&51.2&78.8&68.3&\textbf{64.2}&52.5&\textbf{73.0}&\textbf{90.6}&\textbf{83.4}&{\textbf{68.3}}\\
\midrule
{CoTracker3-Kub~\cite{karaev2024cotracker3}} & 25.39 & \small{$768\times1024$}&47.8&62.0&77.0&72.0&51.9&46.4&67.0&76.2&74.5&{63.9}\\
{CoTracker3~\cite{karaev2024cotracker3}} & {25.39} & \small{$768\times1024$}&49.8&\textbf{64.3}&79.6&\textbf{72.4}&58.4&48.5&71.1&77.9&80.2&{66.9}\\
{AllTracker-Tiny-Kub} & {6.29} & \small{$768\times1024$}&48.4&56.9&78.5&71.2&55.1&49.2&71.1&80.8&78.1&{65.5}\\
{AllTracker-Tiny} & {6.29} & \small{$768\times1024$}&51.6&57.0&79.1&71.1&59.2&50.7&72.4&87.4&79.2&{67.5}\\
{AllTracker-Kub} & {16.48} & \small{$768\times1024$}&51.7&60.2&79.8&71.2&59.1&54.4&71.7&89.2&81.7&{68.8}\\
{AllTracker} & {16.48} & \small{$768\times1024$}&\textbf{53.6}&53.4&\textbf{80.6}&72.3&\textbf{64.3}&\textbf{54.6}&\textbf{73.1}&\textbf{90.6}&\textbf{83.2}&{\textbf{69.5}}\\
\bottomrule
\end{tabular}
\label{tab:highres}
\end{table*}

\section{Experiments}\label{sec:exp}

This section summarizes the key results, which concern the accuracy of our tracker compared to prior state of the art, on a diverse array of test benchmarks. We provide additional details for our experiments in the appendices. 

\paragraph{Metrics and benchmarks}
Our main metric is $\davg$: an accuracy metric with max value 100, capturing how closely the estimated trajectory positions follow the ground truth positions. This is defined as the average of multiple $\delta_k$ metrics, where $\delta_k$ equals $100$ when the estimate is within $k$ pixels of ground truth, measuring at $256 \times 256$ resolution:  
\begin{equation}
    \delta_k = 100 \cdot \mathbbm{1}[ || \mathbf{p} - \mathbf{\hat{p}}||_2 < k ].
\end{equation} 
Averaging $\delta_k$ over $k \in \{1, 2, 4, 8, 16\}$ yields $\davg$. We prioritize this over other considered metrics (e.g., L2~\cite{harley2022particle}, PCK~\cite{harley2022particle}, AJ~\cite{doersch2022tap}) because we find it is both interpretable and robust to outliers. In certain benchmarks we also compute occlusion classification accuracy, and Average Jaccard (AJ) which mixes tracking accuracy with occlusion accuracy.

We evaluate on a total of nine publicly-available benchmarks with point annotations, covering a wide set of domains that include animals (BADJA~\cite{badja}, Horse10~\cite{horse10}), YouTube videos (TAPVid-DAVIS~\cite{doersch2022tap}, TAPVid-Kinetics~\cite{doersch2022tap}), surveillance camera recordings (CroHD~\cite{sundararaman2021tracking}), egocentric recordings (EgoPoints~\cite{egopoints}), and robotics data (RGB-Stacking~\cite{rgbstacking},  RoboTAP~\cite{vecerik2024robotap}). We trim the video lengths to a maximum of 600 frames, and on the larger datasets (CroHD, DriveTrack, EgoPoints, Horse10, Kinetics, RoboTAP) we only track points from the first available query frame, which we find gives similar results to using all possible queries and full video lengths. 
We argue that using a wide array of benchmarks helps flatten the effects of label noise, and gives a better sense of the overall performance in practice. 
We encourage future work to follow our example, but we note that our full evaluation is time-consuming (with some baselines requiring multiple days to finish their pass). 

\paragraph{Baselines} We benchmark a variety of recent point trackers and optical flow models in our tests, including the optical flow models RAFT~\cite{raft} and SEA-RAFT~\cite{wang2024sea}, the long-term flow method AccFlow~\cite{wu2023accflow}, and point trackers PIPs++~\cite{zheng2023point}, LocoTrack~\cite{cho2025local}, DELTA~\cite{ngo2024delta}, BootsTAPIR~\cite{doersch2024bootstap}. We make close comparisons against DELTA~\cite{ngo2024delta} and  CoTracker3~\cite{karaev2024cotracker3}, which are the most recent state-of-the-art point trackers. 
DELTA is relevant because it is a concurrent work focused on dense tracking. 
CoTracker3 is relevant because it outperforms all past work. It has two variants: one trained exclusively on Kubric, and another version finetuned on pseudolabels produced by an ensemble of point trackers on a curated set of 15,000 real videos. This finetuned version of CoTracker3 outperforms all past work.\footnote{The CoTracker3 repo also includes an ``offline'' variant, but it is not consistently better than the standard version, and in any case its memory consumption is outside our compute budget.} We note that the performance of CoTracker3 varies depending on how the query points are grouped~\cite{karaev2023cotracker}. 
The paper suggests to run each query in a separate forward pass, while also adding ``support'' points around the query and a sparse grid covering the image, but in our multi-benchmark evaluation this would be prohibitively expensive (e.g., weeks). We simply give CoTracker3 all queries at once, supplemented by a sparse grid of points around the image. Comparing with values reported by the authors (reproduced in the supplementary), we find that our setup over-estimates CoTracker3's accuracy, but makes our evaluation tractable. 

\paragraph{AllTracker variants} In addition to our main model, which is 16.48 million parameters and trained on a mix of datasets, we evaluate a ``tiny'' version, which is trained similarly to the main model but uses a cheaper CNN backbone and totals only 6.29 million parameters (see details in supplemental). For both versions, we additionally report the results with models trained exclusively on Kubric.

\subsection{Main results}

We evaluate our method's ability to track arbitrary points in diverse videos of various lengths, and compare against the state of the art in Table~\ref{tab:singleres}. For this evaluation we resize all input videos to $384 \times 512$. 
We note that DELTA (the only other dense point tracker) often runs out of memory when testing on a 40G GPU, 
so we use a 96G GPU to evaluate it.

As shown in the table, even our weakest variant, AllTracker-Tiny-Kub, is competitive with state of the art on many datasets, despite being only 6.29M parameters and training on a single synthetic dataset. 
Our full AllTracker model outperforms all other models on average, with the closest competitor being 
the variant of CoTracker3 which relied on an expensive  bootstrapping scheme (noted by ``+15k'' in the table).
This CoTracker3 model wins on three datasets (CroHD, Davis, DriveTrack) and our model wins on the rest and wins on average (66.1 vs. 65.0). 
Our method has a substantial gain over past work on the RGB-Stacking benchmark, which has many query points inside textureless regions; our model's good performance here suggests that it is able to incorporate spatial context from a wider area than the previous models. 
We note that CoTracker3's bootstrapping technique could be combined with our contributions, but it is simpler to gather additional synthetic data as we have done. 

Comparing models trained on Kubric to models trained with a broader data distribution, we see that the additional training data does not reliably improve performance in 
the surveillance dataset CroHD~\cite{sundararaman2021tracking} or the driving dataset DriveTrack~\cite{drivetrack}. 
This suggests room for improvement, in terms of perhaps data balancing or model capacity.

\paragraph{High-resolution performance}
Focusing on the comparison between our model and CoTracker3, we evaluate these models at higher resolutions in Table~\ref{tab:highres}. Here we see that our model's performance reliably increases at higher resolutions, while CoTracker3 plateaus after $448 \times 768$, leading to the surprising result that AllTracker-Tiny outperforms CoTracker3 at $768 \times 1024$. 
Our model is also more memory efficient: with AllTracker we can process videos at $768 \times 1024$ (and produce 786,432 tracks at once) on a 40G A100 GPU, while with CoTracker3, despite only tracking sparse points, we encountered out-of-memory errors until eventually performing the tests on a 96G H100 GPU. 
Our model's relative memory efficiency is partly from using a spatial stride of 8 in the encoder, while CoTracker3 uses stride 4.

\begin{figure}[t]
    \centering    \includegraphics[width=0.9\linewidth]{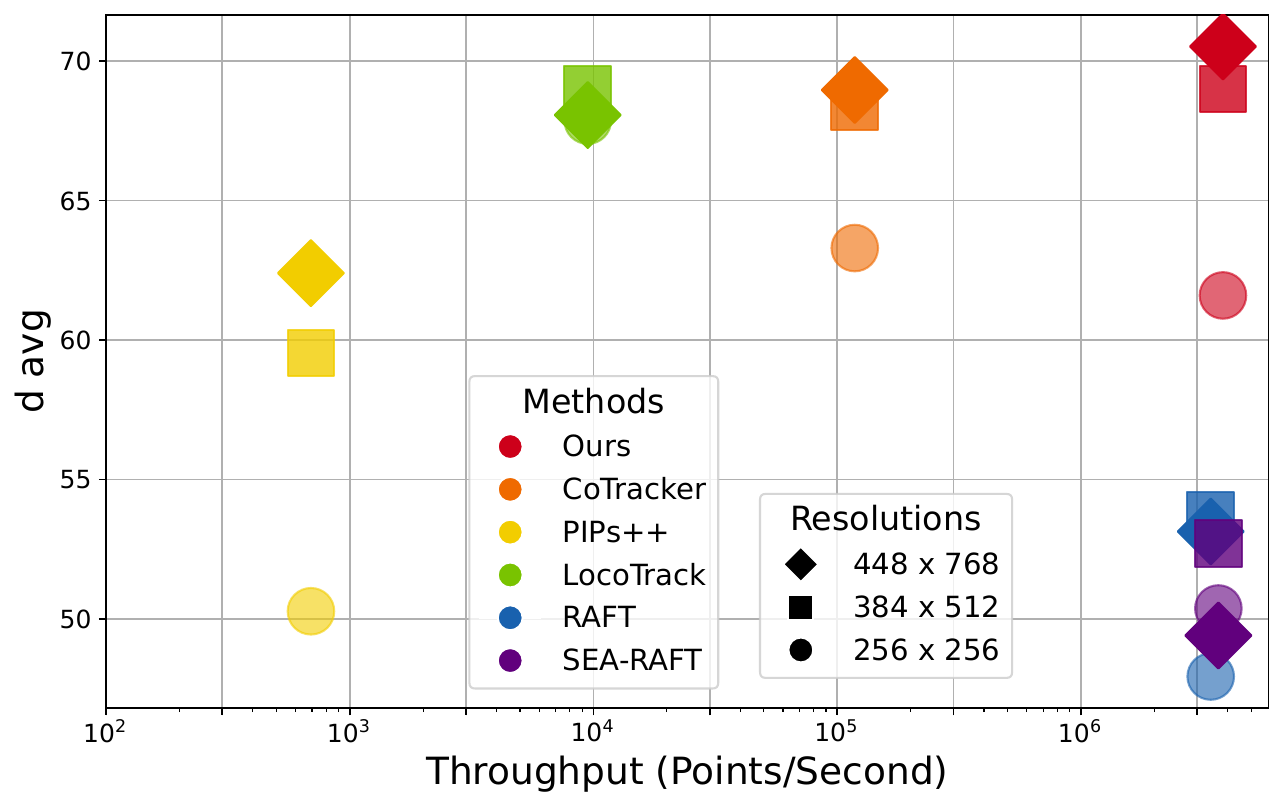}     \vspace{-0.5em}
    \caption{AllTracker (top right corner) delivers accurate multi-frame tracks 
    at the throughput of an optical flow model.}
    \label{fig:throughput}
\end{figure}

\paragraph{Speed} 
We compare the throughput of our model to other point trackers and optical flow models in Figure~\ref{fig:throughput}: our model runs at approximately the speed of optical flow methods, while achieving accuracies higher than the point trackers.

\begin{table}
\small
\centering
\caption{{Average Jaccard evaluation on TAP-Vid datasets.
}\label{tab:aj}} \vspace{-0.5em}
\begin{tabular}{lcccc>{\columncolor[gray]{0.9}}c}
\toprule
Model & Dav. & Kin. & Rgb. & Rob. & Avg. \\
\midrule
CoTracker3 & 62.9&53.5&69.5&66.3&63.1\\
AllTracker & \textbf{63.3}&\textbf{59.1}&\textbf{81.1}&\textbf{71.9}&\textbf{68.9}\\
\bottomrule
\end{tabular}
\vspace{-0.5em}
\end{table}

\begin{table}
\small
\centering
\caption{{Occlusion accuracy evaluation on TAP-Vid datasets.
}\label{tab:oa}} \vspace{-0.5em}
\begin{tabular}{lcccc>{\columncolor[gray]{0.9}}c}
\toprule
Model & Dav. & Kin. & Rgb. & Rob. & Avg. \\
\midrule
CoTracker3 & \textbf{90.1}&85.6&91.6&89.8&89.3\\
AllTracker & \textbf{90.1}&\textbf{90.3}&\textbf{92.8}&\textbf{92.8}&\textbf{91.5}\\
\bottomrule
\end{tabular}
\vspace{-0.5em}
\end{table}

\paragraph{Realtime inference} Our model's sliding-window strategy can be run in ``streaming'' fashion real-time, at a penalty to accuracy: at $512 \times 512$ our model runs at 57.9 FPS and achieves 62.6 $\delta_\text{avg}$ (vs. 66.1 normally); BootsTAPIR~\cite{doersch2024bootstap} has also published a realtime variant, which runs at 21.4 FPS and reaches 62.2 $\delta_\text{avg}$. 

\begin{figure*}[t]
\includegraphics[width=1.0\linewidth]{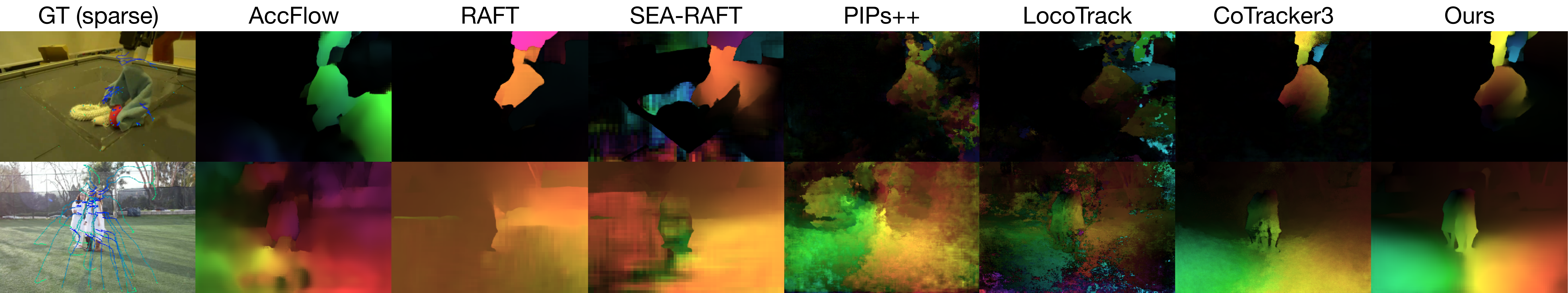}
 \caption{\textbf{AllTracker produces accurate displacement fields across dozens of frames}. 
 Prior optical flow methods struggle to make correspondences across wide time gaps, while our model uses temporal priors to resolve the ambiguity; prior point trackers take multiple minutes to produce output at this density, and show splotchy pattern errors, while our method produces coherent output in less than a second. 
 }
 \label{fig:longrange}
 \vspace{-1em}
\end{figure*}

\begin{table}
\setlength{\tabcolsep}{2pt}
\parbox{.45\linewidth}{
\small
\centering
\caption{A transformer-based temporal module performs better than mixer-based or convolution-based variants.}\label{tab:time_test} \vspace{-0.5em}
\begin{tabular}{lccc}
\toprule
Time component & Acc. \\
\midrule
Transformer~\cite{vaswani2017attention} & \textbf{56.8} \\
MLP-Mixer~\cite{mlpmixer} & 56.1 \\
ConvNext-1D~\cite{liu2022convnet} & 54.6 \\
Conv-1D & 55.8 \\
\bottomrule
\end{tabular}
}
\hfill
\parbox{.5\linewidth}{
\small
\centering
\caption{Representing motion as long-range flow works better than instantaneous 
velocity, and positional embeddings do not help.}\label{tab:motion_test} \vspace{-0.5em}
\begin{tabular}{lc}
\toprule
Motion representation & Acc.\\
\midrule
$0 \rightarrow t$ & \textbf{56.8} \\
$\textrm{emb}(0 \rightarrow t)$ & 54.8\\
$t\!-\!1 \rightarrow t$ & 56.4 \\
$\textrm{emb}(t\!-\!1 \rightarrow t)$ & 55.3\\
\bottomrule
\end{tabular}
}
\vspace{-0.5em}
\end{table}

\begin{table}
\small
\centering
\caption{{Hyperparameter search: It is best to use a ConvNeXt~\cite{liu2022convnet} backbone, 3 refinement blocks, and radius-4 correlations at 5 scales.}\label{tab:hyps}} \vspace{-0.5em}
\begin{tabular}{lcccc}
\toprule
\multirow{2}{*}{Backbone} & Refine & Corr. & Corr. & \multirow{2}{*}{Acc.} \\
  & blocks & radius & scales &  \\
\midrule
\colorbox{white}{ConvNeXt} & 3 & 4 & 5 & \textbf{56.8} \\
\colorbox{highlight}{ConvNeXt w/o pret.} & 3 & 4 & 5 & 54.9 \\
\colorbox{highlight}{BasicEncoder} & 3 & 4 & 5 & 56.1 \\
\colorbox{white}{ConvNeXt} & 3 & \colorbox{highlight}{3} & 5 & 56.5 \\
\colorbox{white}{ConvNeXt} & 3 & \colorbox{highlight}{5} & 5 & 55.5 \\
\colorbox{white}{ConvNeXt} & 3 & {4} & \colorbox{highlight}{4} & 55.5 \\
\colorbox{white}{ConvNeXt} & 3 & {4} & \colorbox{highlight}{6} & Err. \\
\colorbox{white}{ConvNeXt} & \colorbox{highlight}{2} & 4 & 5 & 55.7 \\
\colorbox{white}{ConvNeXt} & \colorbox{highlight}{4} & 4 & 5 & 56.1 \\
\bottomrule
\end{tabular}
\vspace{-0.5em}
\end{table}

\paragraph{Additional metrics} Evaluating AJ and occlusion accuracy on TAPVid datasets, we find that AllTracker obtains substantially better AJ than CoTracker3: 68.9 vs. 63.1 on average (see Table~\ref{tab:aj}, and slightly better occlusion accuracy: 91.5 vs. 89.3 (see Table~\ref{tab:oa}). We perform this evaluation only on the TAP-Vid datasets because we find that other datasets' visibility labels are not as reliable.

\paragraph{Performance on flow benchmarks} While traditional optical flow is not our focus, it is interesting to inspect the model's accuracy in this task. We submitted AllTracker to the official SINTEL test benchmark, yielding end-point error scores of 1.673 on ``clean'' and 3.244 on ``final''; this is not as good as SEA-RAFT~\cite{wang2024sea} (1.309 / 2.601) but comparable to GMFlow~\cite{xu2022gmflow} (1.736 / 2.902). We note that it is common in optical flow literature to produce a different model for each benchmark (finetuned with a particular data mix and resolution), but for this test we simply use our main model as-is. 
Qualitatively, AllTracker's flow maps appear to be coarser than the ones from SEA-RAFT~\cite{wang2024sea}, suggesting that the model is underfitting on this task. 
We provide more optical flow results in the supplementary.

\paragraph{Qualitative results} We visualize the long-range flow maps for our model and selected baselines in Figure~\ref{fig:longrange}. Note that computing the dense flow map for the sparse methods takes multiple minutes, while our method (and the flow methods) produce it in under a second. We find that these flow visualizations reveal differences in the spatial coherence of the tracks: sparse point trackers from the past few years (PIPs++, LocoTrack, CoTracker3) show progressive improvement but include splotchy pattern errors in the motion fields. The optical flow methods (AccFlow, RAFT, SEA-RAFT) show better spatial coherence, but the estimates are unreliable when the displacements are too large. Our method produces motion fields that are both coherent and accurate. 

\subsection{Ablations}

We verify our design choices in an exhaustive series of ablation studies. 
In these experiments, we train each model for 100,000 iterations with 2 GPUs, using Kubric, with 4 inference iterations per sample, at a learning rate of 4e-4, with videos of size $24 \times 384 \times 512$ and $56 \times 256 \times 256$. We test at resolution $384 \times 512$ on a validation dataset that we create from BADJA~\cite{badja}, CroHD~\cite{sundararaman2021tracking}, TAPVid-DAVIS~\cite{doersch2022tap}, DriveTrack~\cite{drivetrack}, Horse10~\cite{horse10}, and RoboTAP~\cite{vecerik2024robotap}, and report the mean $\davg$ across this set. We find that this truncated training and evaluation setup is crucial for enabling a thorough exploration of the model space. We note that the absolute values of these ablation experiments should not be compared to the main experiments, but can be compared to each other.  

\paragraph{Temporal module} Prior work has explored different network architectures for learning tempral priors: PIPs~\cite{harley2022particle} used an MLP-Mixer~\cite{mlpmixer}, PIPs++~\cite{zheng2023point} and TAPIR~\cite{doersch2023tapir} used 1D convolutions (which might reasonably be upgraded to 1D ConvNeXt layers~\cite{liu2022convnet}), and CoTracker~\cite{karaev2023cotracker} used a transformer~\cite{vaswani2017attention}. 
In Table~\ref{tab:time_test} we compare these in our setup and find that a transformer works best. We additionally note that the transformer option is most 
amenable to changing the window size, which is important for jointly training on optical flow (e.g., convolution kernels of size 3 cannot apply to an input of length 2). 

\paragraph{Motion representation} 
Most point trackers use sinusoidal position embeddings of motion, and methods vary on whether displacements should be relative to the query frame or to an adjacent frame. 
Table~\ref{tab:motion_test} shows that displacement relative to the query frame (i.e., frame 0) works best. This choice has the additional benefit of merging the problem with long-range optical flow. 
Table~\ref{tab:motion_test} also shows that sinusoidal embeddings do not help. 

\paragraph{Backbone} Most point trackers use a ``BasicEncoder'' backbone which originated from RAFT~\cite{raft}. We show in Table~\ref{tab:hyps} that a pre-trained ConvNeXt backbone performs better, while a ConvNeXt backbone trained from scratch is worse. 

\paragraph{Hyperparameters} Our main model uses 3 refinement blocks, and radius-4 correlations computed at 5 scales. 
We demonstrate in Table~\ref{tab:hyps} that the neighboring alternatives are worse. We note that using 6 scales produces a runtime error, as $256\times 256$ input is already $1\times 1 $ after 5 scales. 

\paragraph{Huber vs. L1} CoTracker3~\cite{karaev2024cotracker3} and TAPIR~\cite{doersch2023tapir} recommend a Huber loss instead of the L1 used in PIPs~\cite{harley2022particle}; in our setup this is not helpful: 54.9 (Huber) vs. 56.8 (L1).  

\paragraph{Does frame ordering matter?} We train a model with shuffled frames, to disentangle the benefits of joint multi-frame inference from the benefits of temporal continuity: we find that the impact varies from dataset to dataset, but shuffling is worse on average: 56.3 vs. 56.8. 

\section{Conclusion and Limitations} \label{sec:conclusion}

AllTracker is an approach to point tracking that treats the task as multi-frame optical flow. 
Our model delivers state-of-the-art performance on point tracking benchmarks, and produces dense output at high resolution, which past point trackers have struggled to do. 
AllTracker makes sparse point trackers (of similar speed and accuracy) redundant, but interestingly it does not make optical flow methods redundant: it does not outperform the state-of-the-art optical flow methods on optical flow estimation. The model appears to be underfitting on short-range motion estimation, suggesting that better models might be obtained with greater compute. 
A related limitation to address is the temporal window size: with bigger GPUs it may be possible to simply train and test with wider windows~\cite{karaev2024cotracker3}, and resolve longer occlusions. 
A broader area for future work is to add awareness about physical and common-sense constraints on motion, perhaps using 3D~\cite{xiao2024spatialtracker,ngo2024delta} or more expressive model designs~\cite{harley2024tag,zholus2025tapnext}.

\paragraph{Acknowledgments}
This work was supported by the TRI University 2.0 program, ARL grant W911NF-21-2-0104, and a Vannevar Bush Faculty Fellowship.

{
    \small
    \bibliographystyle{ieeenat_fullname}
    \bibliography{99_refs}
}

\clearpage

\noindent{\Large{\textbf{Supplementary Material}}}

\appendix

\section{Additional model details}

\begin{figure}[t]
\centering
\includegraphics[width=1.0\linewidth]{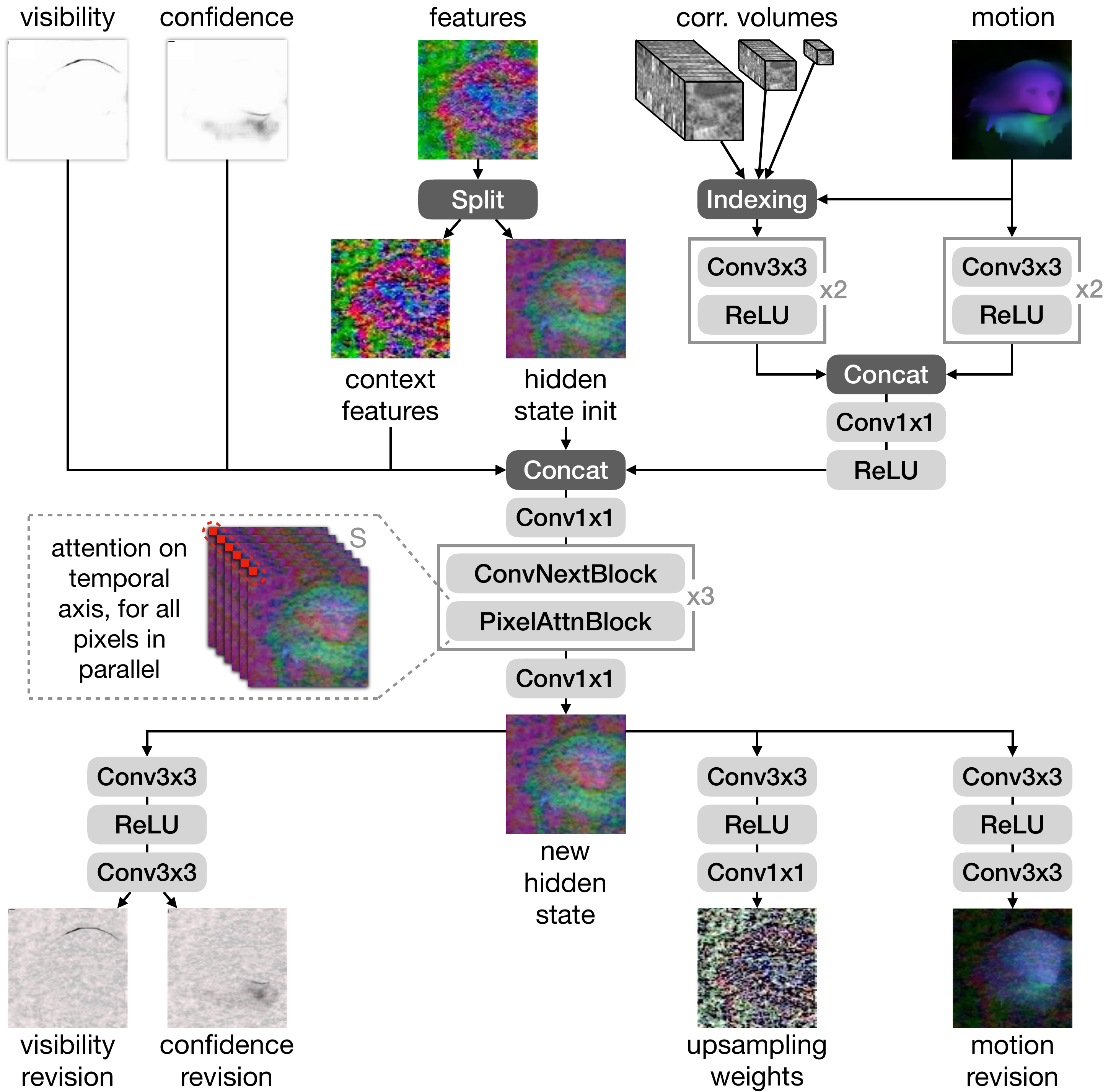}
 \caption{\textbf{Detailed view of iterative refinement block.} We consolidate data from visibility, confidence, correlation, motion, and appearance features into a single feature map, then interleave convolutional spatial blocks and pixel-aligned temporal blocks, and output revisions to the features, visibility, confidence, and motion. 
 This refinement process is iterated 4 times (with shared weights). 
}
 \label{fig:refinement}
\end{figure}

\paragraph{Recurrent module} 
In the recurrent module, we compress and contextualize the input data in stages~\cite{wang2024sea}, following the design ideas of SEA-RAFT~\cite{wang2024sea}, as illustrated in Figure~\ref{fig:refinement}. 
We use parallel 2-layer CNNs (with $3 \times 3$ kernels) on the correlation field and the motion field, then concatenate these feature maps, and merge them with a $1 \times 1$ convolution. We then concatenate the visibility map, confidence map, and appearance features, and merge them to $256$ channels with another $1 \times 1$ convolution. We then apply three ``space-time'' blocks, where the spatial part is a 2D ConvNeXt block  (with a $7\times7$ kernel) and the temporal part is a \textit{pixel-aligned} transformer block (attending the full subsequence span $S$). By pixel-aligned, we mean that attention only happens along the temporal axis (i.e., with cost quadratic in $S=16$), and this is done for every pixel in parallel. 
We note that an important  convenience of our design is that all of the tensors in this stage are aligned with the ``query'' frame; therefore pixel-aligned attention is attention between \textit{corresponding} pixels. 
After the interleaved spatial and temporal blocks propagate tracking-related information across our window, 
we emit a new hidden state, and decode this state into explicit revisions for visibility, confidence, and motion. 
Following SEA-RAFT~\cite{wang2024sea}, we only use half of the feature channels for our recurrent module's hidden state, as shown by the ``split'' step in Figure~\ref{fig:refinement}, which saves memory (reducing the number of output channels from $260$ to $132$) and may also stabilize recurrence.

\paragraph{Model layers} The ConvNeXt blocks are standard: layer-scaled kernel-7 grouped convolution $\rightarrow$ layer norm $\rightarrow$ expansion linear layer (factor 4) $\rightarrow$ GELU $\rightarrow$ reduction linear layer (factor 4) $\rightarrow$ residual add $\rightarrow$ linear layer. The pixel-aligned temporal blocks also have a standard form: layer-scaled transformer block with 8 heads and expansion factor 4 $\rightarrow$ residual add $\rightarrow$ linear layer. 
To inform the attention layers on frame ordering, we add 1D sinusoidal position embeddings to the ``context'' features of a subsequence (see Figure 3 from the main paper), broadcasting these embeddings across the spatial axes. Note that our temporal position embedding is with respect to a subsequence; the model unaware of total video length, and we give no information about the anchor frame's original position in the timeline. 

\paragraph{AllTracker-Tiny} For AllTracker-Tiny, we use a BasicEncoder backbone with channel dimension 128, making only 2.63M parameters (out of 6.29M total), and leave the rest of the architecture unchanged. We use this model's featuremap as both the ``context'' features and the RNN hidden state initialization, rather than splitting a 256-channel featuremap into these two parts.

\begin{table*}[t]
\small 
\centering
\caption{Comparison against CoTracker3 with different evaluation protocols. We evaluate $\delta_\text{avg}$ (higher is better), using an input resolution of $384 \times 512$. The benchmarks are BADJA~\cite{badja}, CroHD~\cite{sundararaman2021tracking}, TAPVid-DAVIS~\cite{doersch2022tap}, DriveTrack~\cite{drivetrack}, EgoPoints~\cite{egopoints}, Horse10~\cite{horse10}, 
TAPVid-Kinetics~\cite{doersch2022tap}, 
RGB-Stacking~\cite{rgbstacking}, and RoboTAP~\cite{vecerik2024robotap}. 
``CoTracker3*'' indicates values reported in the CoTracker3 paper under a more expensive evaluation protocol (and on fewer datasets). Parameter counts are in millions. 
}
\begin{tabular}{lccccccccccc>{\columncolor[gray]{0.9}}c}
\toprule
Method & Params. & Training & Bad. & Cro. & Dav. & Dri. & Ego. & Hor. & Kin. & Rgb. & Rob. & Avg.\\
\midrule
{CoTracker3-Kub~\cite{karaev2024cotracker3}} & 25.39 & Kubric&47.5&\textbf{48.9}&\textbf{77.4}&\textbf{69.8}&58.0&47.5&70.6&83.4&77.2&64.5\\
{CoTracker3*-Kub~\cite{karaev2024cotracker3}} & 25.39 & Kubric&-&-&76.7&-&-&-&66.6&81.9&73.7&-\\
{CoTracker3~\cite{karaev2024cotracker3}} & 25.39 & Kubric+15k&48.3&44.5&77.1&\textbf{69.8}&60.4&47.1&71.8&84.2&81.6&65.0\\
{CoTracker3*~\cite{karaev2024cotracker3}} & 25.39 & Kubric+15k&-&-&76.3&-&-&-&68.5&83.6&78.8&-\\
AllTracker&16.48&Kubric+mix&\textbf{51.5}&44.0&76.3&65.8&\textbf{62.5}&\textbf{49.0}&\textbf{72.3}&\textbf{90.0}&\textbf{83.4}&\textbf{66.1}\\
\bottomrule
\end{tabular}
\label{tab:cotrackerstar}
\end{table*}

\paragraph{Initialization strategy} A close comparison of our model architecture versus SEA-RAFT~\cite{wang2024sea} will reveal that we do not follow SEA-RAFT's choice to directly regress an ``initial'' optical flow estimate with a secondary CNN. Our main reason for omitting this is to save memory. We also note that when frame gaps are sufficiently large, it may in fact be impossible to estimate optical flow without temporal context (e.g., out-of-bounds motion of 200 pixels vs. 300 pixels appears identical), and therefore the flow regression from SEA-RAFT would not likely be as effective as simply propagating the estimates from the previous window.

\begin{figure*}[t]
\includegraphics[width=1.0\linewidth]{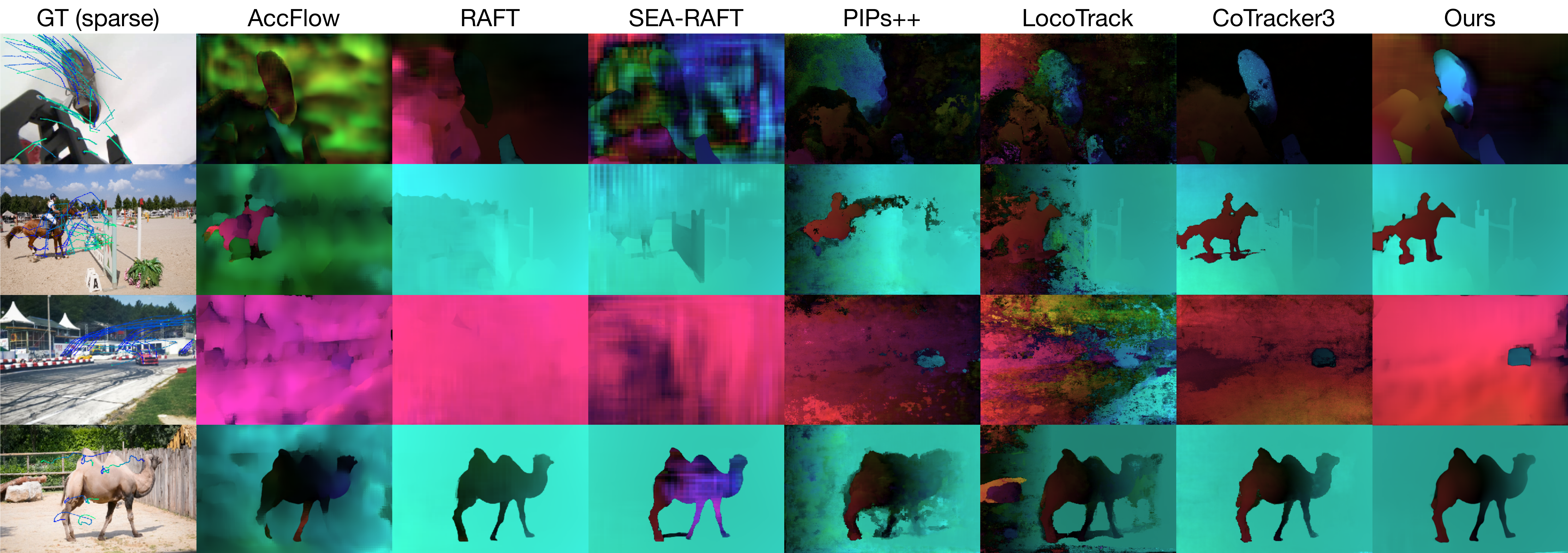}
 \caption{\textbf{Visualization of dense correspondence maps produced by all models.} On the far left column we show the ground truth trajectories overlaid on the first frame of the input video, with  blue-to-green colormap. (Note that a ground truth flow map does not exist in this data.) The flow maps in the other columns show the estimated correspondence field from the first frame of a video to the last frame of the video. Note that RAFT and SEA-RAFT only make use of the first and last frames, while other methods use the intermediate frames as well.}
 \label{fig:denseoutput_alt}
\end{figure*}

\section{Additional training details}

We train with mixed precision in PyTorch (bfloat16). 

From point tracking datasets, we use samples which have anywhere upwards of 256 valid annotated tracks (after augmentations), and trim to a maximum of 6144 tracks to keep memory usage predictable. From optical flow datasets we use dense supervision, but note that this data does not include visibility labels.  

The major variables in speed and memory consumption are batch size, video length, input resolution, and number of refinement steps. To match inference speed across inputs of different length and keep memory consumption within our budget (8x A100 40G), we use the following settings: on optical flow data (where video length is 2), we use batch size 8, resolution $384 \times 768$, and $4$ refinement steps; on videos of length $24$, we use batch size 1, resolution $384 \times 512$, and $4$ refinement steps; on videos of length $56$, we use batch size 1, resolution $256 \times 384$, and $3$ refinement steps. In the first stage of training (on Kubric alone), we use only videos of length 24 and 56, and split our 8 GPUs by using 4 for the 24-frame videos and 4 for 56-frame videos. In the second stage of training (on the wider mix of data), we split our 8 GPUs  by using 1 for optical flow, 3 for videos of length 24, and 4 for videos of length 56. We sync gradients across GPUs after each backward pass.

For our BCE loss, we apply the sigmoid first and then use the direct BCE loss, which we (counter-intuitively) found to be more numerically stable than BCE with logits.  

We note that \textit{no architecture modifications are required} to train jointly for optical flow estimation and point tracking. In optical flow, the temporal attention is a redundant operation, but we do not disable the temporal transformer, as there are still MLP layers within it which participate in the processing.

\section{Additional baseline details}

As mentioned in the main paper, the performance of CoTracker-style models 
depends how the query points are grouped. Intuitively, if multiple queries lie on the same object, they will be tracked more accurately. When these methods are tasked with tracking all of the benchmark queries at once, they tend to exploit a bias in the data and perform better than they would perform on random queries. The authors of these methods suggest a strategy for mitigating this effect, which consists of running each query in a separate pass, while also adding ``support'' points around the query and a sparse grid covering the image. In our high-resolution multi-benchmark evaluation, these steps would be prohibitively expensive (e.g., weeks). We therefore simply give the models the advantage of the data bias: we give all queries at once, and supplement them with a sparse grid of points around the image. We compare this evaluation protocol to author-reported results in Table~\ref{tab:cotrackerstar}. 
We find that our cheaper protocol over-estimates the accuracy of CoTracker3, but our own model is still more accurate on average.

\section{Additional optical flow results}

\begin{table}
\small
\centering
\caption{{Optical flow end-point error (``EPE-All'') 
in the offical SINTEL test benchmark. \vspace{-0.5em}
}\label{tab:sintel}}
\begin{tabular}{lcc}
\toprule
Model & Clean & Final \\
\midrule
SEA-RAFT~\cite{wang2024sea} & \textbf{1.309} &  \textbf{2.601} \\
RAFT~\cite{raft} & 1.609 & 2.855  \\
GMFlow~\cite{xu2022gmflow} & 1.736 & 2.902 \\
PWC-Net~\cite{sun2018pwc} & 4.386 & 5.042	\\
AllTracker & 1.673 & 3.244 \\
\bottomrule
\end{tabular}
\end{table}

\begin{table}
\small
\centering
\caption{{Optical flow end-point error in the CVO ``Final'' (T=7) and ``Extended'' (T=48) test sets, for visible/occluded pixels. \vspace{-0.5em}}\label{tab:cvo}}
\begin{tabular}{lcc}
\toprule
Model & T=7 & T=48 \\
\midrule
AccFlow~\cite{wu2023accflow}     & 1.15 / 4.63 & 28.1 / 52.9\\
DOT~\cite{le2024dense}         & \textbf{0.84} / \textbf{4.05} & 3.71 / \textbf{7.58}\\
AllTracker  & 1.03 / 4.10 & \textbf{3.41} / 7.93\\
\bottomrule
\end{tabular}
\end{table}

We ran AllTracker on the official SINTEL test benchmark, yielding the scores shown in Table~\ref{tab:sintel}, with other official scores included for comparison. We note that it is common in optical flow literature to produce a different model for each benchmark (finetuned with a particular data mix and resolution), but we 
simply use our original checkpoint. AllTracker's optical flow is not as accurate as SEA-RAFT, but comparable to RAFT or GMFlow. 
Qualitatively, AllTracker's flow maps appear to be coarser than the ones from SEA-RAFT~\cite{wang2024sea}, suggesting that the model is underfitting; better models might be obtained with greater compute. 

We additionally evaluate in CVO, the multi-frame optical flow dataset used by AccFlow and DOT, showing results in Table~\ref{tab:cvo}. We find that on short sequences DOT (combining CoTracker2 and RAFT) performs best; on long sequences AllTracker performs best. We also find that AllTracker is 3x the speed of DOT.

In sum, 
these results suggest that AllTracker is not state-of-the-art for optical flow estimation, even though it includes optical flow data in its training. Attaining top performance optical flow and point tracking with a single model remains an open challenge.

\section{Additional ablation details}

\paragraph{Validation dataset} As mentioned in the main paper, we construct a validation dataset for our ablation studies, using BADJA~\cite{badja}, CroHD~\cite{sundararaman2021tracking}, TAPVid-Davis~\cite{doersch2022tap}, DriveTrack~\cite{drivetrack}, Horse10~\cite{horse10}, and RoboTAP~\cite{vecerik2024robotap}. The purpose of these studies is to obtain a quick (but reliable) look at performance, and therefore we do not use these datasets in their entirety, and note we also exclude some of our available datasets. 
We subsample from these datasets by (1) selecting the first frame with {any} annotations and tracking only the queries on that frame, (2) trimming all videos to a maximum length of 300 frames. These choices, along with our truncated training regime (training only 100,000 steps and only on Kubric) allows for most ablation experiments to (individually) start and finish within 24 hours.

\begin{figure}[t]
    \centering
    \includegraphics[trim=0 5 0 0, clip, width=0.7\linewidth]{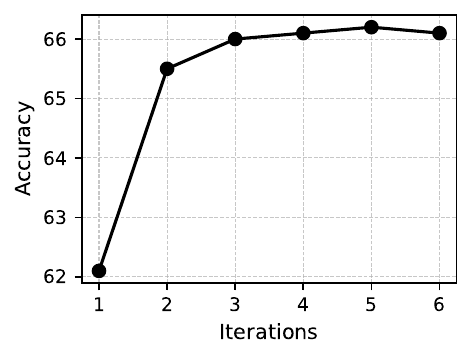}
     \vspace{-1em} 
    \caption{Accuracy over inference steps. Accuracy rises quickly then plateaus. In the main evaluation we use 4 iterations.}
    \label{fig:acc_over_iters}
\vspace{-1em}
\end{figure}

\paragraph{Inference steps} In the main paper we apply the recurrent refinement module 4 times. Figure~\ref{fig:acc_over_iters} shows performance at different iterations, evaluating $\delta_\textrm{avg}$ over all datasets and averaging, using an input resolution of $384 \times 512$. On the first step the model achieves 62.1 accuracy, which already outperforms most state-of-the-art models. Accuracy rises to its peak at 5 iterations, then begins to drop. In the main paper we report results at 4 iterations, because we find that step 4 and step 5 produce similar accuracy at higher resolutions. 

\section{Additional qualitative results}

To obtain long-range flow estimates from our point tracker baselines, we query them to track every pixel of the first frame of the video. We perform these queries in ``batches'' of 10,000, which is the maximum that fits on our GPU. Note that CoTracker3 benefits from processing these jointly, whereas PIPs++ and LocoTrack do not, due to the design of these models. 
To obtain long-range flow estimates from the optical flow baselines, we pair the first frame with every other frame, creating $T-1$ frame pairs for flow estimation, where $T$ is the length of the video. 

We show additional visualizations of multiple models' dense outputs in  Figure~\ref{fig:denseoutput_alt}. We notice striking dissimilarity across the outputs of the methods, attesting to the difficulty of the task, and to the usefulness of visualizing point tracks as flow maps. 
We find that when the foreground displacements are large, the flow models often ``give up'' on the dynamic foreground and produce a motion field that only describes the background. We also find that PIPs++ and LocoTrack often struggle with spatial smoothness, while the flow models do not. CoTracker3 occasionally fails on smoothness too (see row 3 with the car), but less so. Our model appears to produce results that are smooth and accurate, which matches intuitions for a model that blends the 2D processing of flow models with the temporal coherence of point trackers.

\end{document}